\documentclass[runningheads]{llncs}
\usepackage{adjustbox}
\usepackage{multirow}
 
\usepackage{eccv}
\usepackage{svg}
\usepackage{pifont}
\usepackage{booktabs}

\usepackage{caption}
\usepackage{eccvabbrv}

\usepackage{graphicx}
\usepackage{booktabs}
\usepackage{wrapfig}

\usepackage[accsupp]{axessibility}  

\usepackage[pagebackref,breaklinks,colorlinks,citecolor=eccvblue]{hyperref}
\usepackage{hyperref}[colorlinks,linkcolor=blue]
\usepackage{orcidlink}
\usepackage{bbding}

\begin{document}


\title{Harmonizing Light and Darkness: A Symphony of Prior-guided Data Synthesis and Adaptive Focus for Nighttime Flare Removal}


\titlerunning{Harmonizing Light and Darkness}


\author{Lishen Qu\inst{1,2}\and
Shihao Zhou\inst{1,2} \and
Jinshan Pan\inst{3} \and
Jinglei Shi\inst{1} \and \\
Duosheng Chen\inst{1}\and
Jufeng Yang\inst{1,2} 
}
\authorrunning{Qu et al.}

\institute{VCIP \& TMCC \& DISSec, College of Computer Science, Nankai University \and
Nankai International Advanced Research Institute (SHENZHEN· FUTIAN) \and
School of Computer Science and Engineering, Nanjing University of Science and Technology}



\maketitle

\begin{abstract}


    %
    Intense light sources often produce flares in captured images at night, which deteriorates the visual quality and negatively affects downstream applications. In order to train an effective flare removal network, a reliable dataset is essential. The mainstream flare removal datasets are semi-synthetic to reduce human labour, but these datasets do not cover typical scenarios involving multiple scattering flares. To tackle this issue, we synthesize a prior-guided dataset named Flare7K*, which contains multi-flare images where the brightness of flares adheres to the \textit{laws of illumination}. Besides, flares tend to occupy localized regions of the image but existing networks perform flare removal on the entire image and sometimes modify clean areas incorrectly. Therefore, we propose a plug-and-play Adaptive Focus Module (AFM) that can adaptively mask the clean background areas and assist models in focusing on the regions severely affected by flares. Extensive experiments demonstrate that our data synthesis method can better simulate real-world scenes and several models equipped with AFM achieve state-of-the-art performance on the real-world test dataset. Code is available at \href{https://github.com/qulishen/Harmonizing-Light-and-Darkness}{https://github.com/qulishen/Harmonizing-Light-and-Darkness}.

\end{abstract}

\addtocounter{figure}{-2}
\begin{figure}[t]
    \begin{minipage}{0.58\textwidth}
    \includegraphics[width=1\textwidth]{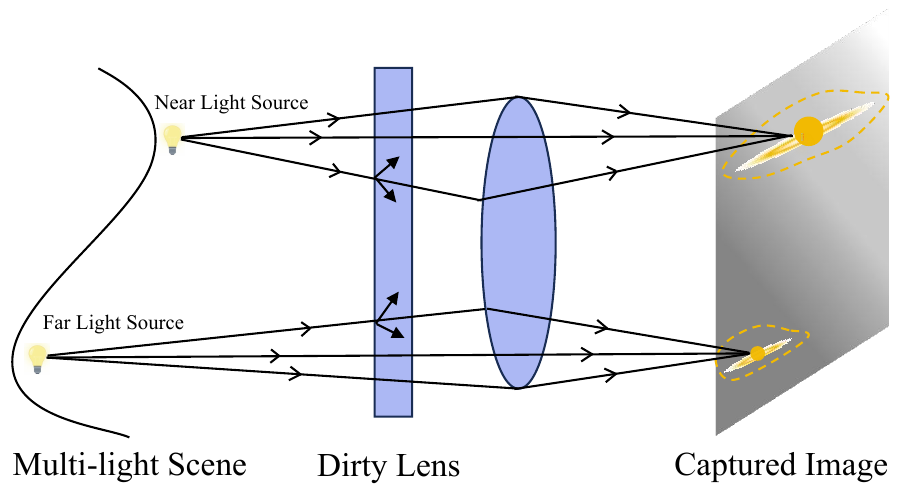}
    \captionsetup{labelformat=empty}
    \vspace{-0.7cm}
    \caption{(a) The schematic diagram}
    \end{minipage}
    \hspace{2mm}
    \begin{minipage}{0.38\textwidth}
    \centering
    \includegraphics[width=0.36\textwidth,height=0.36\textwidth]{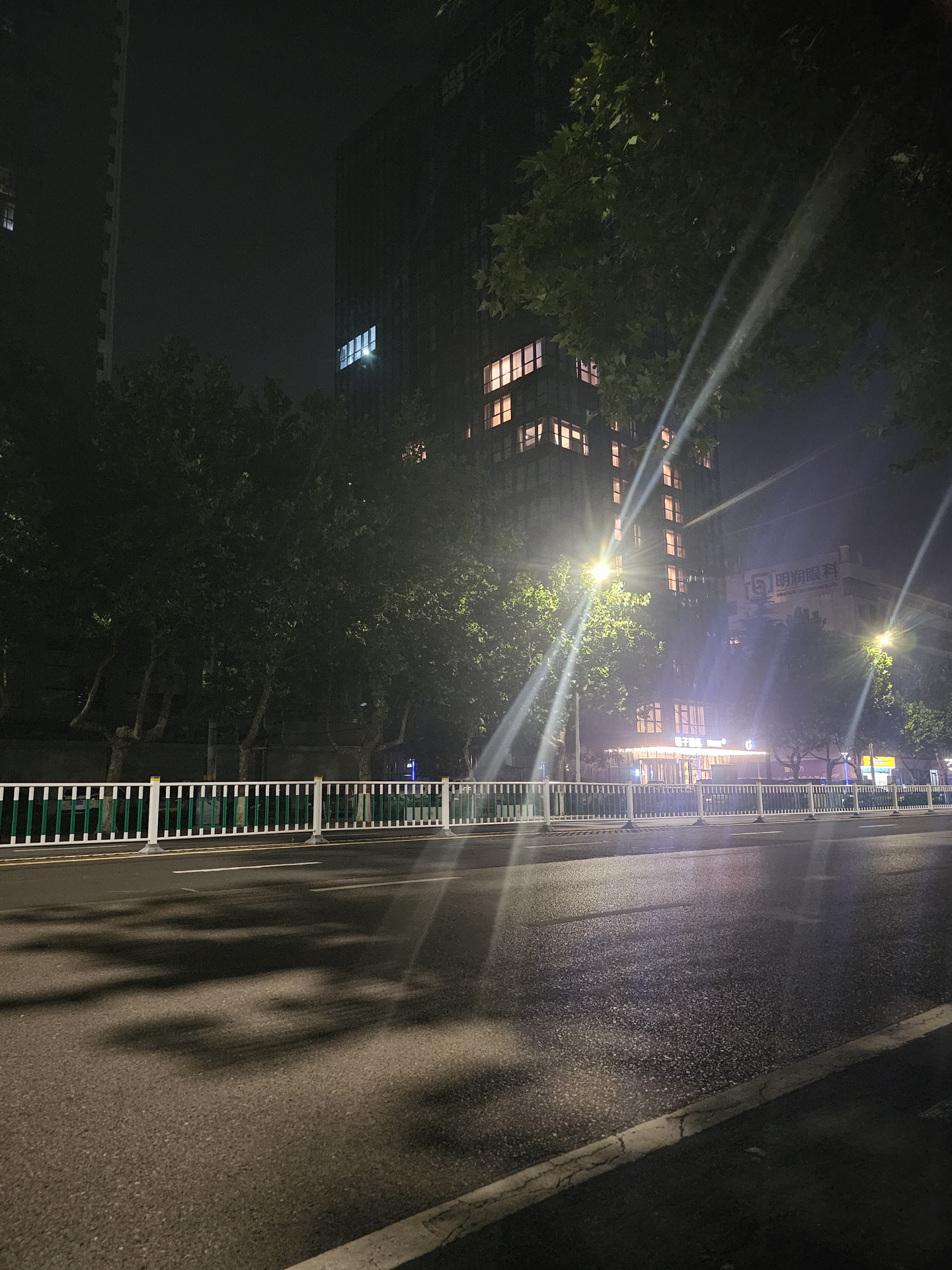}
    \includegraphics[width=0.36\textwidth,height=0.36\textwidth]{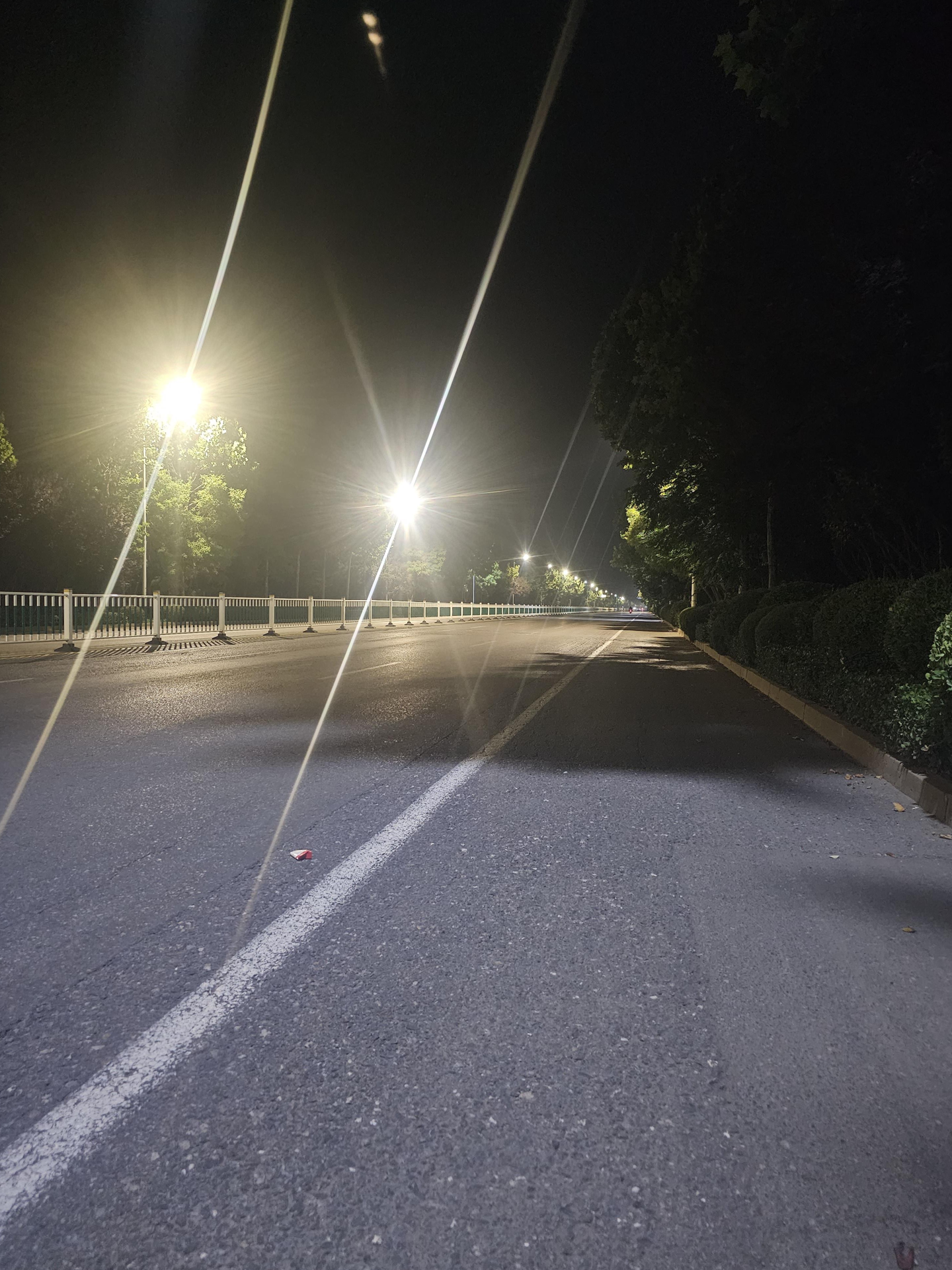}
    
    \includegraphics[width=0.36\textwidth,height=0.36\textwidth]{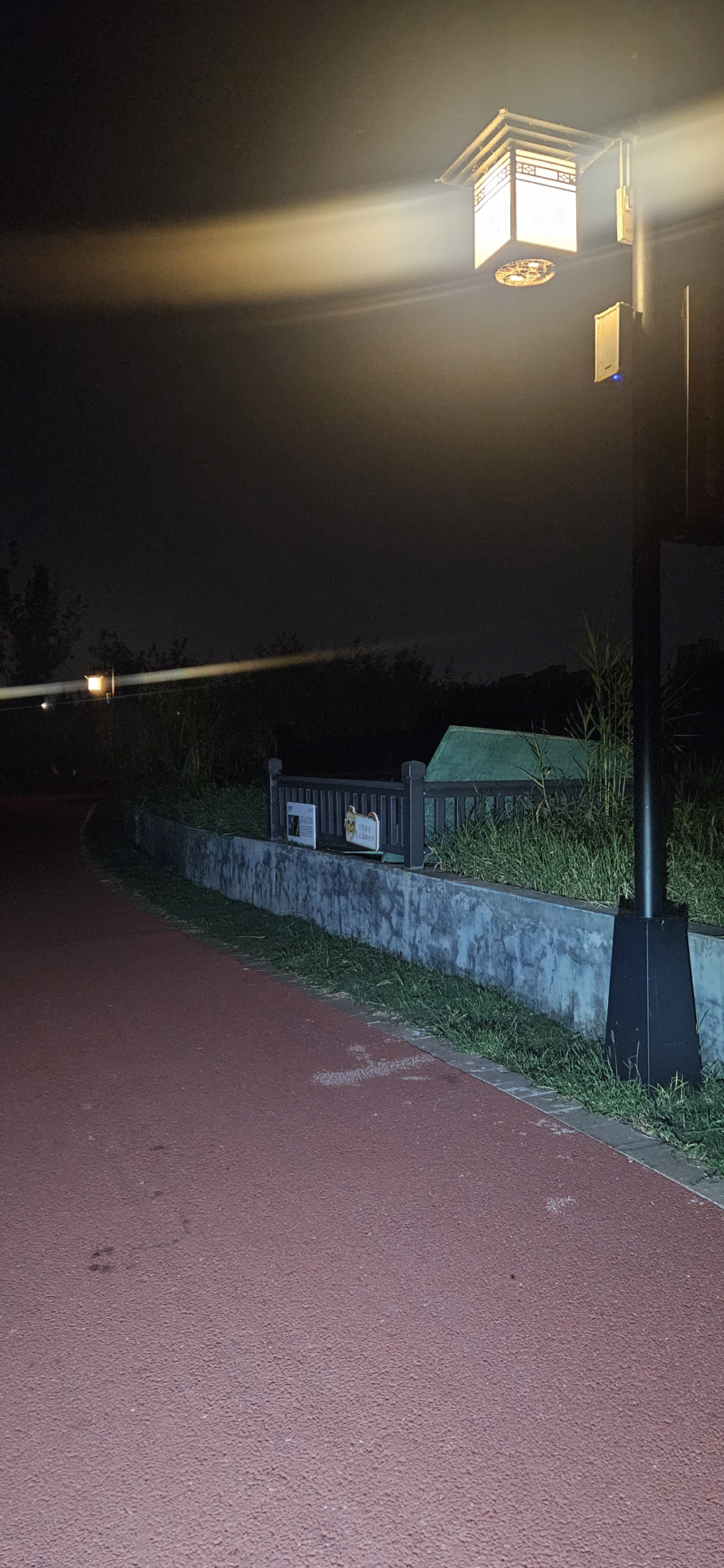}
    \includegraphics[width=0.36\textwidth,height=0.36\textwidth]{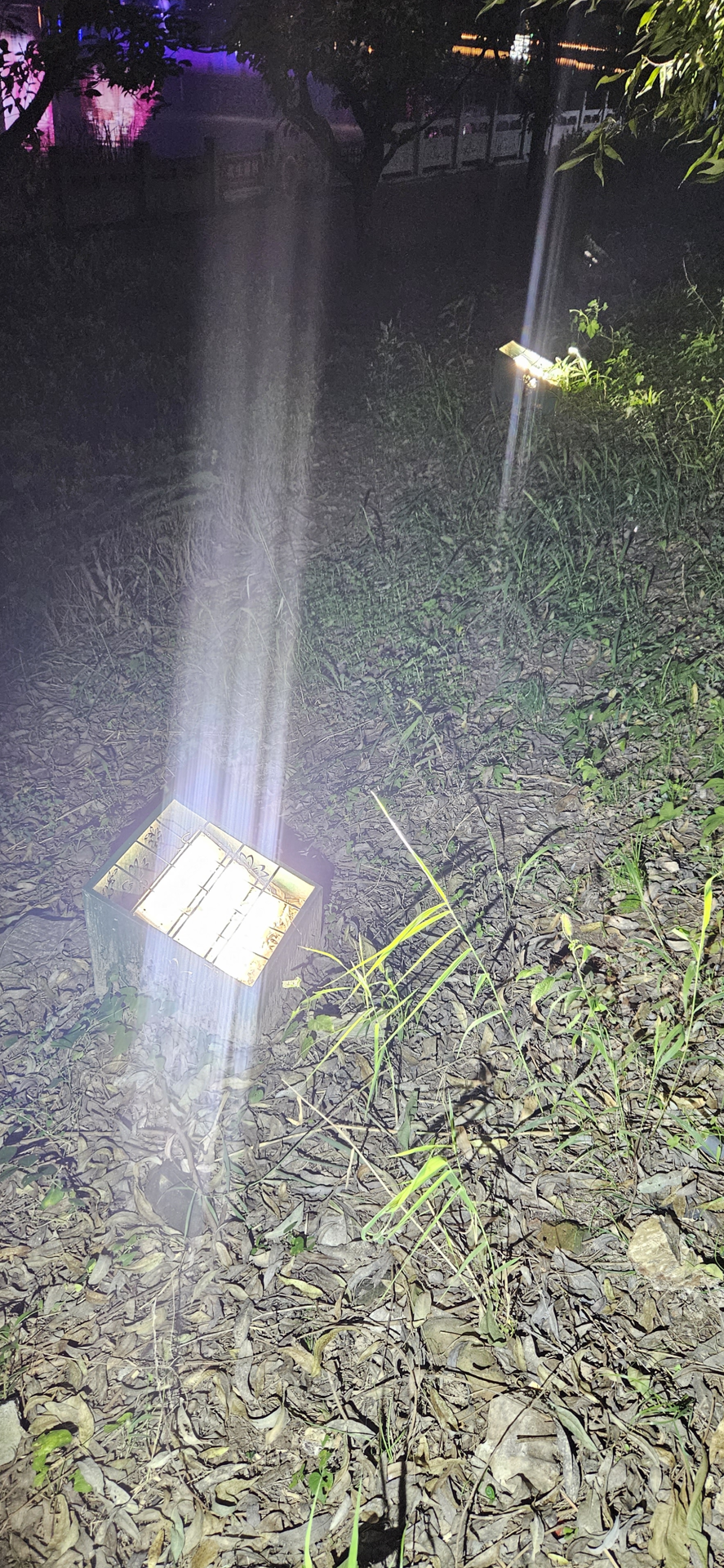}
    \captionsetup{labelformat=empty}
    \caption{(b) Real-world captured images}
    \end{minipage}
    \caption{In most realistic scenarios, flares caused by multiple artificial light sources usually show different brightness. Intuitively, the brightness of flares decreases as the distances of light sources increase. These flares typically occupy localized regions (indicated by the yellow dashed areas) on the captured image.}
    \label{fig:hebing}
\end{figure}

\section{Introduction}
\label{sec:intro}
It can be frustratingly difficult to prevent camera lenses from being affected by grease, scratches, dust, and other particles.
Light scatters on the surface of these lenses, resulting in radial line patterns (\ie, scattering flares) on captured images.
%
Even when lenses are clean, we occasionally find some ghosting (\ie, reflective flares), which are caused by reflections at the air-glass interface of the lens.\cite{hullin2011physically}.
These phenomena, especially at night, not only affect the perception quality of visual content like videos and photographs but also have adverse effects on downstream visual applications, such as nighttime driving with stereo cameras\cite{dai2022flare7k} and target tracking\cite{nussberger2015robust}.
%
\par
A common approach\cite{coating1,coating2} is to employ anti-reflective (AR) coatings on the lens aimed at reducing internal reflections. However, this method is prohibitively expensive for consumer-grade cameras and offers limited effectiveness.
%
As image restoration algorithms continue to evolve, using deep learning methods to remove nighttime flares becomes more convenient and effective than employing AR coatings.
%
%
%
%
%
%
\par
%
%
However, training deep learning models requires a large dataset and collecting a real-world dataset containing flare-corrupted images is a laborious task.
%
To address this issue, Wu \etal\cite{wu2021train} introduce the first semi-synthetic flare removal dataset, and offer a training pipeline for flare removal.
Based on the work of Wu \etal, Dai \etal\cite{dai2022flare7k} create a larger and more realistic flare dataset called Flare7K, including 5,000 scattering flares and 2,000 reflective flares, which serves as a benchmark for nighttime flare removal. 
%
%
\par
%
%
Since the flare images and the background images are separated in the semi-synthetic datasets mentioned above, the method used to synthesize the flare-corrupted images is also very important. Therefore, many researchers are working on creating a methodologically sound data synthesis pipeline.
Zhou \etal\cite{zhou2023improving} synthesize flare-corrupted images using the convex combination to avoid distribution shift and overflow.
Dai \etal\cite{dai2023nighttime} propose a new optical centre symmetry prior to synthesize a more realistic reflective flare removal dataset.
These efforts aim to synthesize more realistic datasets but struggle to effectively address the challenge of multi-flare images which are more commonly seen than single-flare images.
%
Besides, as shown in \cref{fig:hebing}, it is no coincidence that the brightness of flares generated by different light sources is related to their distance.
The laws of illumination describe the relationship among the illumination on a plane, the distance of the light source, and the angle of incidence of light, allowing us to synthesize multi-flare images that adhere to objective rules.
Thus, we propose a prior-guided data synthesis method that utilizes the laws of illumination to obtain a more realistic dataset named Flare7K*. 
\par
%
We further note that the nighttime flares tend to occupy local regions in the image and these regions are commonly brighter than the clean background, as shown in \cref{fig:hebing}. 
However, existing flare removal methods typically affect the entire image, which consequently leads to unnecessary or incorrect adjustments.
Besides, the low-light areas of nighttime images often contain high-frequency noise that can interfere with useful information.
Therefore, we propose an Adaptive Focus Module (AFM), which can adaptively mask some low-light regions and help models focus on the flare regions.
%
%
Trained on Flare7K*, several models equipped with AFM outperform current state-of-the-art (SOTA) methods, which demonstrates the effectiveness of our method. 
%
\par
Our contributions can be summarized as follows:

\begin{itemize}
\item[$\bullet$] We synthesize a physically realistic dataset named Flare7K* based on the laws of illumination, which can address the limitations of existing synthesis methods in generating datasets with an adequate variety of scenes.
\item[$\bullet$] We introduce a plug-and-play Adaptive Focus Module (AFM), which not only helps models achieve better performance in removing flares but also avoids touching clean regions.

\item[$\bullet$] Extensive experiments demonstrate the benefits of Flare7K* for training a superior model, and several baselines equipped with AFM show performance improvements.

\end{itemize}

\section{Related Work}

\subsection{Flare Removal Datasets}
To train a neural network for flare removal, a diverse and authentic flare dataset is crucial.
As previously mentioned, mainstream datasets primarily consist of semi-synthetic data due to the labour-intensive and repetitive nature of collecting real-world flare data pairs.
Wu \etal \cite{wu2021train} introduced the first semi-synthetic flare dataset, which, despite its valuable contributions, exhibits different characteristics from real-world flares and achieved limited performance.
Last year, Dai \etal proposed two datasets, namely Flare7K\cite{dai2022flare7k} and Flare7K++\cite{dai2023flare7k++}, which have become crucial benchmarks for nighttime flare removal.
Recently, the Nighttime Flare Removal competition at MIPI 2024 (the Mobile Intelligent Photography and Imaging Workshop 2024) also provided a dataset that includes 600 pairs of 2k resolution images, serving as a supplement to Flare7K and Flare7K++.
Besides, a Raw Image Dataset\cite{phone_data} has emerged that is specifically tailored for mobile photography.
\par
Apart from datasets, a methodologically sound data synthesis approach is also crucial, and there have been many attempts to synthesize more realistic datasets.
%
Unlike the previous methods\cite{dai2022flare7k,wu2021train} of adding a flare image directly to a background image, Zhou \etal blended the scene layer and flare layer using a convex combination to prevent flare images from experiencing distribution shift and overflow.
Jin \etal \cite{similar} theoretically proved the similarity of scattering flares which points out that multiple scattering flares often exhibit similar shapes.
However, these data synthesis methods have limitations, notably the inability to synthesize multi-flare images and ensure that flare brightness adheres to physical laws.
We hence utilize the prior knowledge of the law of illumination to build a physically realistic synthesis pipeline.
\par
\subsection{The Frameworks in Flare Removal}
Dai \etal\cite{dai2022flare7k} validated the superiority of Flare7K against Uformer \cite{Uformer}, Restormer \cite{Zamir2021Restormer}, MPRNet \cite{MPRnet}, U-Net \cite{U-net}, and HINet \cite{chen2021hinet}, with Uformer achieving the best performance.
%
%
Since the light source may be removed during the flare removal process in these networks, Dai \etal pasted the light source back in post-processing.
However, this operation requires manually selecting an appropriate threshold and accurately restoring small light sources can be challenging.
To get better preservation of light sources, FF-Former\cite{zhang2023ff} implemented a Light Source Mask Loss Function which can decrease the burden of the network.
In Dai \etal's new work\cite{dai2023flare7k++}, the pipeline was modified to preserve the light source in the ground truth of the training dataset and the output of Uformer was increased from 3 channels to 6 channels.
\par
A two-stage architecture\cite{kotpflare}, combining Uformer\cite{Uformer} and Dense-Vision-Transformer (DPT)\cite{dpt} to simultaneously obtain depth and pixel information, is also employed.
While our method also utilizes deep information, it is only used for synthesizing the training dataset and does not impose an additional computational burden on the network.

\subsection{Task-Specific Prior}
Priors refer to patterns and knowledge that are challenging to acquire through model training but can be manually obtained from past experiences and theoretical reasoning\cite{priorall}.
There are many priors, such as the Additive Composite Model (ACM)\cite{rain0} used in image deraining\cite{rain1,rain2,rain3}, the gradient guidance prior adopted in super-resolution\cite{SR1,SR2}, and the Retinex model\cite{low-light0} for low-light enhancement\cite{low-light1,low-light2,low-light3}.
The most relevant task to nighttime flare removal is nighttime haze removal\cite{nighttimehaze1,nighttimhaze2,nighttimhaze3}, as the glare effect caused by fog at night shares a similar shape with a flare.
In the realm of deep image dehazing, the atmospheric scattering model (ASM)\cite{daqi} is commonly employed, as evidenced by its integration into many approaches \cite{dehazenet,dehaze1,dehaze2}. 
Some other image dehazing algorithms\cite{dehaze_dark1,dehaze_dark2,dehaze_dark3} are implemented based on the Dark Channel Prior \cite{dark_prior}.
Given the distinct physical principles underlying lens flares and the glare effect induced by fog, it is valuable to explore specific prior knowledge for lens flare removal.
%
%
\par
Dai \etal proposed a new optical centre symmetry prior \cite{dai2023nighttime} based on the centrosymmetric relationship between reflective flares and light sources. 
We observe a correlation between the brightness of multiple flares and the distance of the light source from the lens: the closer the light source, the brighter the flare produced.
Besides, flares in nighttime images often occupy specific regions of the image, appearing brighter than the other clean regions.
Therefore, guided by these priors, we develop a novel data synthesis method and an Adaptive Focus Module (AFM) for nighttime flare removal.





\begin{figure}[t]
    \centering
    \begin{subfigure}{1\textwidth}
        \centering
        \includegraphics[width=0.95\textwidth]{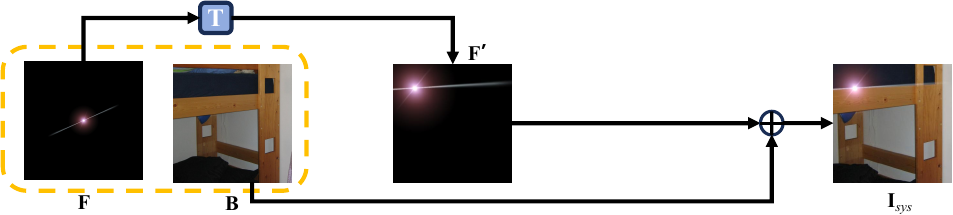}
        \caption{The existing data synthesis method\cite{dai2022flare7k,dai2023flare7k++}}
        \label{fig:img1}
    \end{subfigure}
    \qquad
    \begin{subfigure}{1\textwidth}
        \centering
        \includegraphics[width=0.95\textwidth]{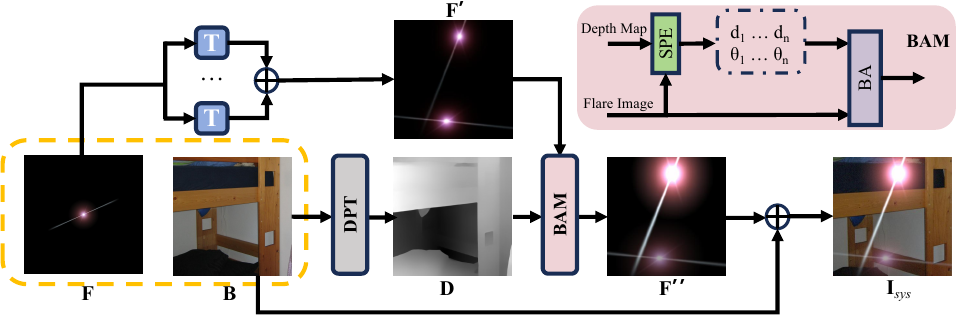}
        \caption{Our data synthesis method}
        \label{fig:img2}
    \end{subfigure}
    \qquad
    \begin{subfigure}{1\textwidth}
        \centering
        \includegraphics[width=0.95\textwidth]{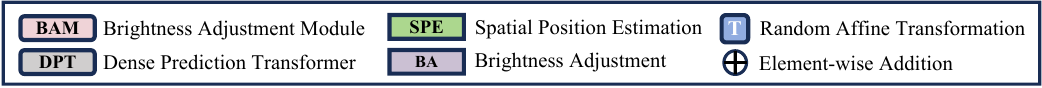}
    \end{subfigure}
    \caption{Figure (a) shows the previous method for synthesizing flare images. Figure (b) is our proposed data synthesis pipeline. Our synthesis method can effectively simulate scenes with multiple flares of varying brightness.}
    \label{fig:sysmodel}
\end{figure}


\section{Proposed Prior-guided Data Synthesis Approach}
\subsubsection{Overview of Synthesis Pipeline.}
\label{sec:operation1}
As existing data synthesis methods are limited in simulating scenarios with multiple flares, which hinders the model's ability to effectively remove flares in such situations, we propose a prior-guided data synthesis method for synthesizing the physically realistic dataset called Flare7K*. 
Different from existing methods, our data synthesis method can synthesize multi-flare images while ensuring their brightness adheres to physical laws.
The comparison between our synthesis method and the previous method is depicted in \cref{fig:sysmodel}.
\par
First, we perform multiple random affine transformations of the same flare separately to get several flares.
Due to the need for depth information to adjust the brightness of flares, we estimate the depth map of the background image using a pre-trained Dense Prediction Transformer (DPT) model\cite{dpt}.
%
Then, we develop a Brightness Adjustment Module (BAM), which will be detailed in \cref{sec:bam}, to adjust the brightness of multiple flares that have been affine transformed.
The above operations can be expressed by:
\begin{equation}
\mathbf{F_{i}''}  =\mathbf{BAM}(\mathcal{T}_{i}(\mathbf{F} ), \mathcal{D} (\mathbf{B} ) )=\mathbf{BAM}(\mathbf{F_{i}'},\mathbf{D}),
\end{equation}
where $\mathcal{T}$ and $\mathcal{D}$ denote the Random Affine Transformation and Depth Estimation, respectively; $\mathbf{F}$ denotes the origin flare image; $\mathbf{F_{i}}'$ is the flare image after the $i^{th}$ random affine transformation; $\mathbf{F_{i}}''$ is the flare image after brightness adjustment; $\mathbf{B}$ represents the background image and \textbf{D} represents its depth map.
\par
Finally, we capture multiple flares after adjusting the brightness, then add them to the background image to obtain the final flare-corrupted image.
The final flare-corrupted image can be represented as:
\begin{equation}
    \mathbf{I}_{\mathrm{sys}}  = Clip(\mathbf{B} + \sum_{i=1}^{n} \mathbf{F_{i}''} ),
    \label{equal:hecheng}
\end{equation}
\noindent where $Clip(\cdot)$  denotes clipping the addition to the range of $\left [0 ,1 \right ] $, and $n$ represents the total number of flares to be generated.

\par
In our work, we use \cref{equal:hecheng} to regulate the brightness of multiple flares. 
The Flare7K* dataset synthesized by our pipeline can simulate a greater variety of scenes, with the flares on the images conforming to objective laws.
The subsequent experiments will demonstrate that Flare7K* can enhance the model's adaptability in real-world captured images, particularly those with multiple flares of varying brightness.
\subsubsection{Brightness Adjustment Module (BAM).}
\label{sec:bam}
%
Intuitively, the farther the light source is from the lens, the less bright the flare appears.
The laws of illumination allow for a quantitative portrayal of this physical phenomenon and the formula is as follows:
\begin{equation}
E = \frac{I\cos\theta  }{d^{2} },
\label{equal:law}
\end{equation}

\noindent where $E$ indicates the illumination at a point on the plane, and $I$ is the luminous intensity of the light source. 
$\theta$ is the angle between the optical axis and the incident light rays, and $d$ is the distance from the light source to the illuminated point. 
For the sake of explanation simplicity, we abuse the terms "angle of incidence " and "depth" in the remainder of the paper.
\par
Since multiple flares are transformed from one flare, only the angle of incidence of light rays and depth of different light sources need to be calculated.
We first perform the Spatial Position Estimation (SPE) using the depth map and affine-transformed flare images to get the depth and the angle of incidence. SPE operation can be expressed by the following equation:
\begin{equation}
d_{i},\theta_{i}   = \mathcal{S} \space  (\mathbf{F_{i}'},\mathbf{D}),
\label{equal:spatial}
\end{equation}
\noindent where $\mathcal{S}(\cdot)$ denotes the operation of SPE.
\par
In SPE, we utilize the average depth of all pixel points of the light source as the actual distance between that light source and the lens, as expressed by the following equation:
\par
\begin{equation}
    d_{i} = \frac{1}{N} \sum_{j=1}^{N} \mathbf{D}_{(x_{j},y_{j})} \space \space ,\space (x_{j},y_{j}) \in \mathbf{L_{i}'},
    \label{equal:depth}
\end{equation}
where $N$ represents the total number of pixel points in the depth map and $\mathbf{L_{i}}'$ symbolizes the light source area of the affine-transformed flare. $x_{j}$ and $y_{j}$ denotes the position of the $j^{th}$ pixel point.
\par
For the calculation of the incident angle, since the camera parameters used for taking photos are unknown, we use the horizontal field of view to estimate.
According to the law of similar triangles, we can obtain the following formula:
\begin{equation}
\theta_{i} = \arctan (\frac{2r_{i}}{W}  \cdot \tan \frac{\varphi }{2}),
\label{equal:theta}
\end{equation}
where $\varphi$ represents the horizontal field of view,  $W$ denotes the width of the background image, and $r_{i}$ denotes the average distance from the pixel points of the ${i^{th}}$ light source to the centre of the image.
\par
Compared to the scenes being captured, the size of the lens can be ignored and represented as a point. 
After obtaining the depth and the incident angle, we substitute these values into \cref{equal:law} to calculate the illumination of the lens from different light sources.
\par
By using the above equations, we can adjust the brightness of each flare. 
The formula for making the final brightness adjustment is as follows:
\par
\begin{equation}
\mathbf{F''_{i}} = \mathbf{F'_{i}} \cdot  \frac{E_{i}}{\frac{1}{n}\cdot\sum_{i=1}^{n} E_{i} }  = \mathbf{F'_{i}} \cdot (\frac{\bar{d}}{d_{i}})^2\cdot\cos \theta_{i},
\end{equation} 
where $\bar{d}$ is the average depth of all light sources.
Specifically, we use the light with a $0^{\circ}$ incident angle and $\bar{d}$ as a reference for adjusting the brightness of each flare. 
Last but not least, due to variations in the fields of view among different cameras, training various models with the same dataset may lead to poor robustness.
Compared to the previous fixed synthesis method, our method can synthesize a dataset for a specific camera by adjusting the field of view $\varphi$. 
The subsequent experiments will compare the differences in quantitative results when choosing different fields of view.
%
    
\par

\section{Proposed Adaptive Focus Module}
Mainstream flare removal methods typically remove flares across the entire image, inadvertently affecting clean areas as well.
%
%
%
As mentioned before, nighttime flares tend to occupy a localized area in the image and exhibit greater brightness compared to the background. 
%
Moreover, the low-light regions of nighttime images often contain significant amounts of irrelevant high-frequency noise, which can severely impede model training.
%
%
\par
 \begin{figure}[t]
     \centering
     \includegraphics[width=1\linewidth]{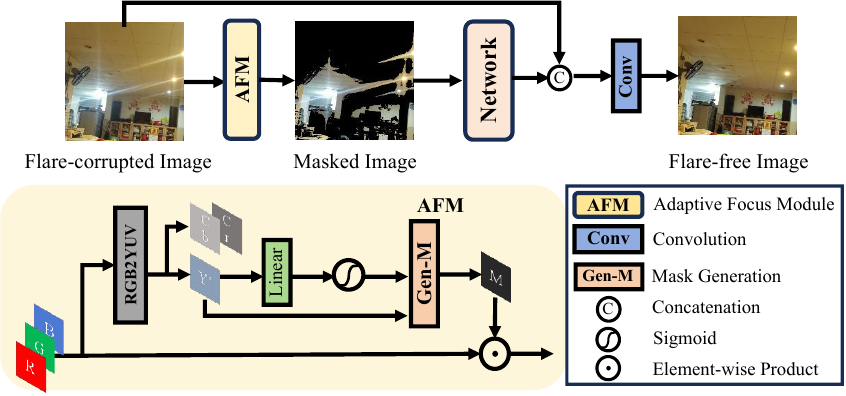}
     \caption{Overview of our training pipeline. The main purpose of AFM is to obtain the image that only contains the flare region and the network can be any commonly used image restoration network.}
     \label{fig:mask}
     \vspace{-3mm }
 \end{figure}
Based on the above analysis and to address the limitation of existing networks, we develop a plug-and-play Adaptive Focus Module (AFM), which can help existing models adaptively focus on the flare regions and improve their performance.
The structure of our training pipeline, which incorporates AFM, is depicted in \cref{fig:mask}.
\par
To enhance the distinction between dark and bright areas and minimize the parameters of AFM, we use prior knowledge of the relationship between RGB values and luminance.
We convert the image from RGB formats to $\mathbf{Y}'$CbCr formats defined by ITU-R BT.601.
Subsequently, the $\mathbf{Y}'$ channel which represents the luminance matrix is sequentially processed through a linear layer and a sigmoid function to obtain a luminance threshold, as shown in \cref{equal:threshold}:

\begin{equation}
    \tau = \sigma (Linear(\mathbf{Y}')),
    \label{equal:threshold}
\end{equation}
where $\sigma(\cdot)$ represents the sigmoid function and $\tau$ denotes the luminance threshold.
%
The mask $\mathbf{M}$, which is of the same size as the original image but with only one channel, is generated by the following formula:
\begin{equation}
    \mathbf{M}_{(i,j)} = \begin{cases}
1 , \mathbf{Y}'_{(i,j)} \ge \tau\\ 
0 , \mathbf{Y}'_{(i,j)} <  \tau
\end{cases},
\end{equation}
\noindent and the masked image $\mathbf{I} _{\mathrm{M}}$ is expressed as shown in the following equation:
\begin{equation}
    \mathbf{I} _{\mathrm{M}}= \mathbf{M}  \odot \mathbf{I} _{\mathrm{input}},
\end{equation}
where $\odot$ represents the element-wise multiplication. After the above operations, the masked image $\mathbf{I} _{\mathrm{M}}$ is fed into the network for training. 
Compared to the original input $\mathbf{I} _{\mathrm{input}}$, $\mathbf{I} _{\mathrm{M}}$ carries less irrelevant information and the model will not inadvertently touch clean background areas.
Subsequently, we concatenate the original image with the predicted result of the localized region from the network.
Finally, the concatenated outcome is passed through a 1 $\times$ 1 convolution layer to restore the original background details.
The comprehensive process is delineated by the subsequent formula:
\begin{equation}
\mathbf{I}_{\mathrm{output}} = Conv (Concat (Net(\mathbf{AFM} (\mathbf{I}_{\mathrm{input}})) ,\mathbf{I}_{\mathrm{input}} )),
\end{equation}
where $Net(\cdot)$ denotes the operation performed by a commonly used flare removal network. $\mathbf{I}_{\mathrm{input}}$ and $\mathbf{I}_{\mathrm{output}}$ represent the flare-corrupted image and the predicted flare-free image.
\par
AFM reduces the amount of irrelevant information in the image before it enters the network, helping the model focus on the region of flare degradation and avoid affecting the clean background.
The most important aspect is that AFM can be readily integrated into several baseline models, assisting them in improving their performance.
Subsequent experiments verify its effectiveness quantitatively and qualitatively.
\section{Experiments}
\subsection{Experiments Settings}

\subsubsection{Evaluation Metrics.}
We employ PSNR, SSIM\cite{wang2004image}, and LPIPS\cite{LPIPS} to measure the image restoration quality. Additionally, we adopt G-PSNR and S-PSNR, as introduced by Dai \etal\cite{dai2023flare7k++}, which respectively denote the PSNR of the glare region and the PSNR of the streak region. In the experiment results, the best and second-best scores are \textbf{highlighted} and \underline{underlined}.
%
\par

\vspace{-2mm}
\subsubsection{Training Details.}
Since Uformer\cite{Uformer} performed best in the previous work\cite{dai2022flare7k,dai2023flare7k++}, we also train a Uformer using the same optimizer and learning rate to ensure a fair comparison.
%
In the work of Dai \etal\cite{dai2023flare7k++}, Uformer, Restormer\cite{Zamir2021Restormer}, MPRNet\cite{MPRnet} and HINet\cite{chen2021hinet} were used. We also train these baseline models equipped with AFM on the dataset synthesized by our method.
For the models that exceed GPU memory, we adopt the same parameter reduction approach as previous works\cite{dai2022flare7k,dai2023flare7k++}. 
The loss functions in our work align with the previous works\cite{dai2022flare7k,dai2023flare7k++,kotpflare}, comprising the $L_{1}$ loss, the perceptual loss with a pre-trained VGG-19\cite{vggloss} and the reconstruction loss.

\vspace{-2mm}

\subsubsection{Data Aggregation Approach.}
In addition to adding multiple flares and the brightness adjustment operation in \cref{sec:operation1}, we use the same data aggregation approach as in previous works\cite{dai2022flare7k,zhang2023ff}. 
We conduct six experiments on the selection of the field of view, selecting $20$, $40$, $60$, $80$, $100$, and random degrees, respectively.
Besides, regarding the operation of adding multiple flares, we randomly select flare numbers ranging from $1$ to $3$ for addition.
%

\subsection{Comparison to Other Methods}

\par
\begin{figure*}[t]
\scriptsize
\centering
\begin{tabular}{cccccc}
\begin{adjustbox}{valign=t}
\begin{tabular}{cccccc} 
\includegraphics[width=0.16\textwidth]{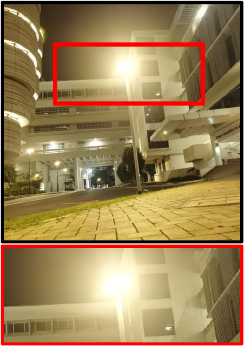}
\includegraphics[width=0.16\textwidth]{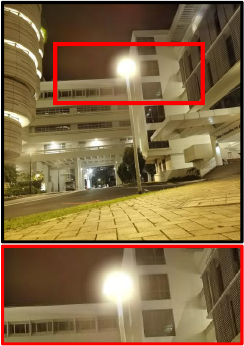}
\includegraphics[width=0.16\textwidth]{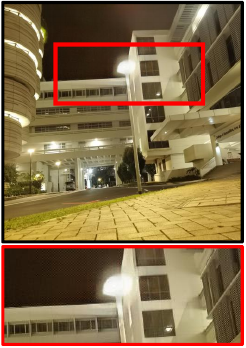}
\includegraphics[width=0.16\textwidth]{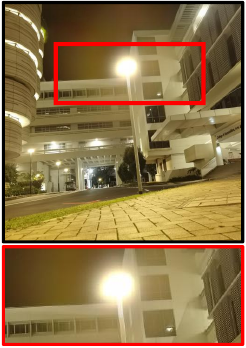}
\includegraphics[width=0.16\textwidth]{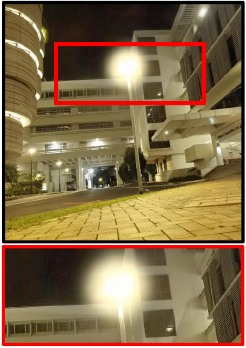}
\includegraphics[width=0.16\textwidth]{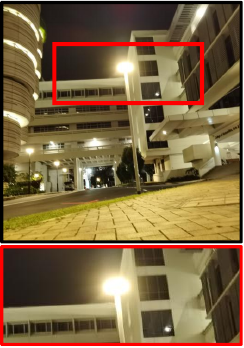}
\\
\includegraphics[width=0.16\textwidth]{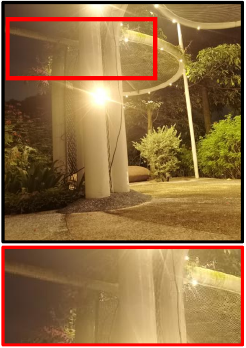}
\includegraphics[width=0.16\textwidth]{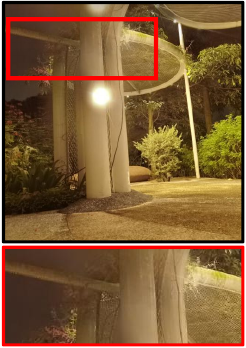}
\includegraphics[width=0.16\textwidth]{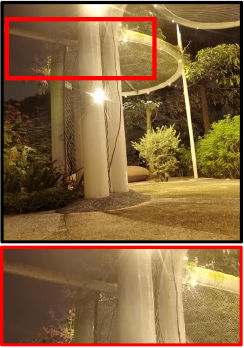}
\includegraphics[width=0.16\textwidth]{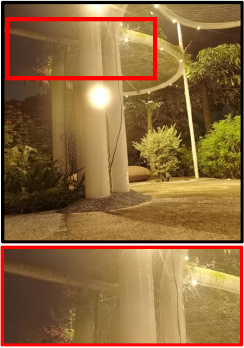}
\includegraphics[width=0.16\textwidth]{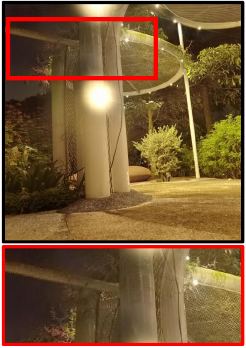}
\includegraphics[width=0.16\textwidth]{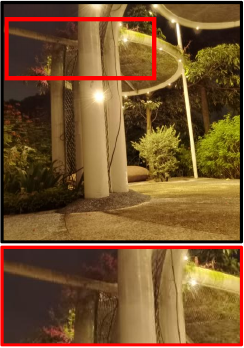}
\\
\includegraphics[width=0.16\textwidth]{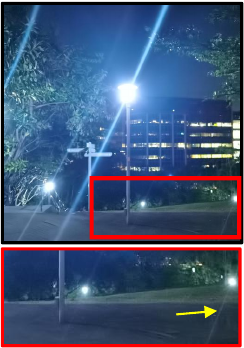}
\includegraphics[width=0.16\textwidth]{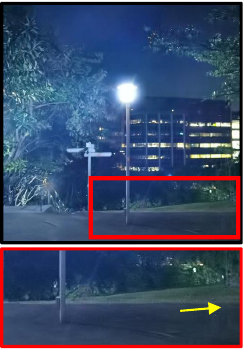}
\includegraphics[width=0.16\textwidth]{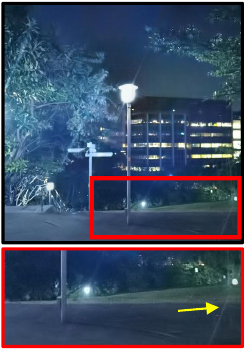}
\includegraphics[width=0.16\textwidth]{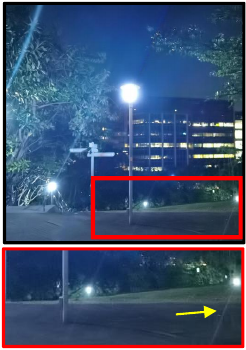}
\includegraphics[width=0.16\textwidth]{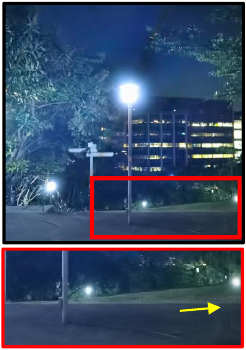}
\includegraphics[width=0.16\textwidth]{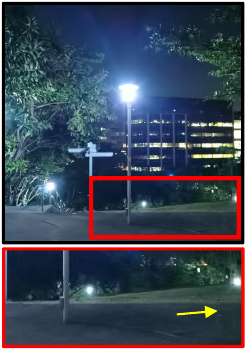}
\\
\includegraphics[width=0.16\textwidth]{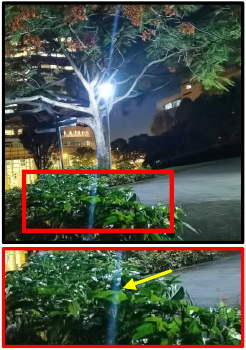}
\includegraphics[width=0.16\textwidth]{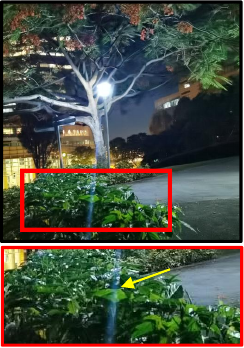}
\includegraphics[width=0.16\textwidth]{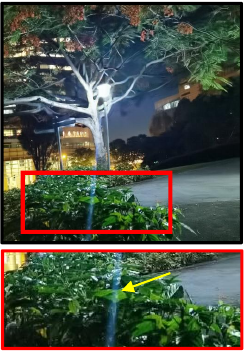}
\includegraphics[width=0.16\textwidth]{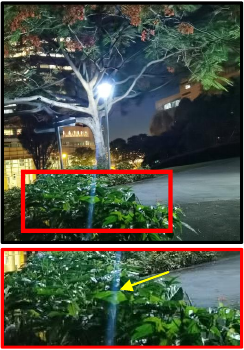}
\includegraphics[width=0.16\textwidth]{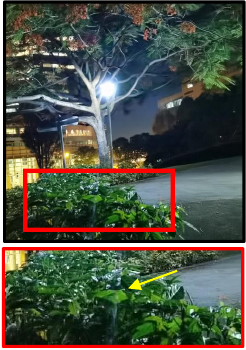}
\includegraphics[width=0.16\textwidth]{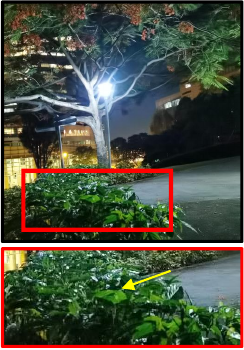}
\\

\hspace{-5mm}
(a) Input  \hspace{7mm}
(b) Dai\cite{dai2022flare7k}  \hspace{5mm}
(c) Zhou \cite{zhou2023improving}\hspace{3mm}
(d) Dai++\cite{dai2023flare7k++} \hspace{4mm}
(e) Ours\hspace{9mm}
(f) GT 
\end{tabular}
\end{adjustbox}
\end{tabular}
\caption{Qualitative comparisons with the state-of-the-art methods on the real-world test dataset. Dai denotes the first Flare7K\cite{dai2022flare7k} work of his team and Dai++ denotes their second Flare7K++\cite{dai2023flare7k++} work.}
\label{fig:comparison}
\vspace{-3mm}
\end{figure*}

\subsubsection{Qualitative comparison.}

%
We carry out visual quality evaluation and compare previous methods\cite{wu2021train,zhou2023improving,dai2022flare7k,dai2023flare7k++} with ours, as shown in \cref{fig:comparison}.
The previous methods have some flaws.
The method of Zhou \etal\cite{zhou2023improving} may result in overexposure of the light source, as in the first, second and third rows.
Dai \etal's method\cite{dai2022flare7k} results in the incorrect removal of the light source, as shown in the second image of the third row.
From the first and second row of \cref{fig:comparison}, our method can eliminate the foggy flare than the others and preserve the light source better.
Upon analyzing this phenomenon, we discover that the previous methods may cause the model to misinterpret foggy flares as light sources, leading to the failure to remove them. 
From the third and fourth rows of \cref{fig:comparison}, it can be observed that our method is more effective in removing weak flares than the other methods.
The primary reason for this is that the previous dataset does not include images with multiple flares of varying brightness, a capability that Flare7K* possesses.
The comparisons and analyses demonstrate that our method achieves better visual quality results than SOTA.

\subsubsection{Quantitative comparison.}
To further demonstrate the effectiveness of our proposed method, we conduct quantitative experiments to compare previous methods with ours.
%
%
Apart from the three methods in the visual comparison, earlier methods such as the nighttime dehazing method proposed by Zhang \etal\cite{zhang2018single} and the nighttime visual enhancement method proposed by Sharma \etal\cite{nighttime_sharma} can also be used for removing flares.
%
%
%
%
In terms of quantitative results, our method achieves the best performance, as shown in Table~\ref{tab:Flare7K}. 
\par
After training Uformer on Flare7K* and applying the AFM technique, our method achieves an improvement of approximately $0.512$ dB in PSNR compared to the best method, noting that the improvement represents a significant enhancement in flare removal.
Besides, the LPIPS decreases by approximately 8\% compared to the previous best method, indicating that our images have better perceptual quality.
All metrics are better than SOTA which demonstrates the effectiveness of our proposed methods. 
To avoid the specificity of Uformer, in subsequent experiments, we will validate the effectiveness of our method on multiple baselines.
%
\par

\begin{table}[t]
\vspace{-1mm}
\caption{
Quantitative comparison with the state-of-the-art methods on the real-world test dataset. It is worth noting that since most of the results in previous methods do not use the Flare-R training set in Flare7K++, ours also do not use it to ensure a fair comparison.
}
\label{tab:Flare7K} 
\centering
\resizebox{1.0\textwidth}{!}{
\begin{tabular}{l@{\hspace{0.3cm}}cccccccc}
\toprule
\multirow{2}{*}{{Method}} & \multirow{1}{*}{}  & \multirow{1}{*}{Zhang~\textit{et~al.}} &\multirow{1}{*}{Sharma~\textit{et~al.}} &\multirow{1}{*}{Wu~\textit{et~al.}} &\multirow{1}{*}{Dai~\textit{et~al.}} &\multirow{1}{*}{Zhou~\textit{et~al.}} &\multirow{1}{*}{Dai~\textit{et~al.}}& 
 \hspace{2mm} \multirow{2}{*}{\textbf{Ours}}
\\ 
\multirow{1}{*}{} & \multirow{1}{*}{}  & \multirow{1}{*}{\cite{nighttime_zhang}} &\multirow{1}{*}{\cite{nighttime_sharma}} &\multirow{1}{*}{\cite{wu2021train}} &\multirow{1}{*}{\cite{dai2022flare7k}} &\multirow{1}{*}{\cite{zhou2023improving}} &\multirow{1}{*}{\cite{dai2023flare7k++}}
\\ \midrule
PSNR$\uparrow$                                &                                   & 21.022                                               & 20.492                                                 & 24.613                                   & 26.978       &25.184      &\underline{27.257}        &  \hspace{2mm} \textbf{27.769}                                                                  \\ 
SSIM$\uparrow$                                &                                    & 0.784                                               & 0.826                                                 & 0.871                                   & \underline{0.890}      & 0.872     &\underline{0.890}    &  \hspace{2mm} \textbf{0.895}                                                                       \\ 
LPIPS$\downarrow$                             &                                   & 0.1738                                               & 0.1115                                                 & 0.0598                                   & \underline{0.0466}     &0.0548   & 0.0471     &  \hspace{2mm} \textbf{0.0429}                                                                            \\ 
G-PSNR$\uparrow$                              &                                   & 19.868                                                    & 17.790                                                      &  21.772                                      & 23.507   &22.112   &\underline{23.762}        &  \hspace{2mm} \textbf{24.010}                                                                              \\ 

S-PSNR$\uparrow$                              &                                  &     13.062                                                &   12.648                                                    &      16.728                                   & \underline{21.563}       &20.543
     &21.294       &  \hspace{2mm} \textbf{22.849}             \\ \bottomrule
\end{tabular}
}
\vspace{-3mm}
\end{table}

\vspace{-5mm}
\subsection{Ablation Study}
\subsubsection{Impact of Flare7K*.} 
\begin{figure*}[t]
\scriptsize
\centering
\begin{tabular}{cc}
\begin{adjustbox}{valign=t}
\begin{tabular}{cc} 
\includegraphics[width=0.46\textwidth]{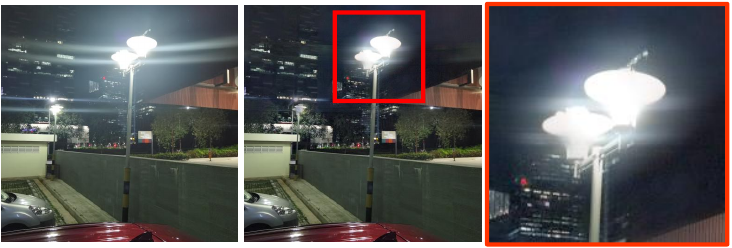}
\hspace{5mm}
\includegraphics[width=0.46\textwidth]{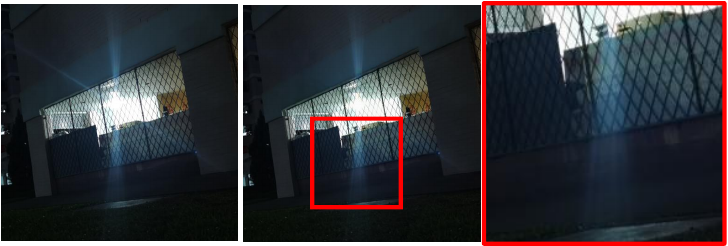}
\\
\hspace{-18mm}
(a) Input \hspace{4mm}
(b) Flare7K\cite{dai2022flare7k} \hspace{27mm}
(a) Input \hspace{4mm}
(b) Flare7K\cite{dai2022flare7k}
\\
\includegraphics[width=0.46\textwidth]{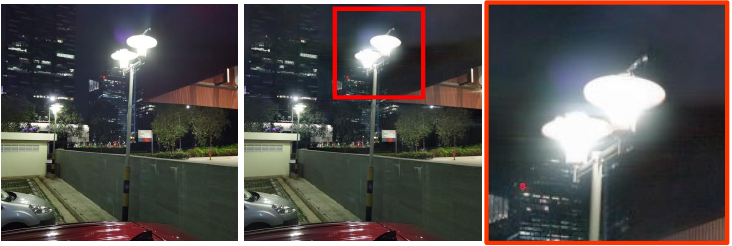}
\hspace{5mm}
\includegraphics[width=0.46\textwidth]{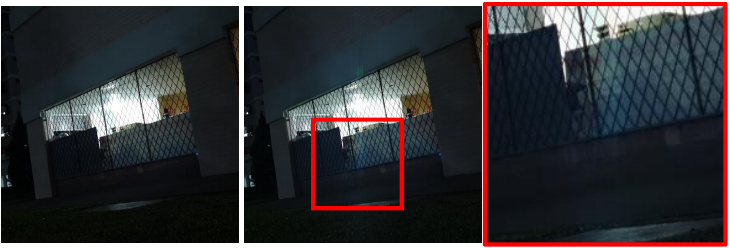}
\\
\hspace{-20mm}
(c) Reference   \hspace{2mm}
(d) Flare7K*      \hspace{27mm}
(c) Reference   \hspace{2mm}
(d) Flare7K*
\end{tabular}
\end{adjustbox}
\end{tabular}
\caption{Visual comparison of the ablation experiment on Flare7K*.}
\label{fig:data_sys}
\end{figure*}

To focus on exploring the visualization enhancements facilitated by Flare7K*, we compare the inference results of Uformer trained on Flare7K and Flare7K*.
%
On the one hand, the model trained on Flare7K may struggle to remove flares near light sources, but the same model trained on Flare7K* removes them more thoroughly, as shown in the images on the left of \cref{fig:data_sys}.
On the other hand, compared with other methods in \cref{fig:comparison}, our method is more effective in removing weak flares, which can be attributed to our Flare7K* dataset, as shown in the images on the right of \cref{fig:data_sys}.
\par
To quantitatively validate the effectiveness of our data synthesis method, we conducted an ablation study, as shown in \cref{tab:data-sys}. Additionally, to ensure the completeness of our experiment, we conduct an experiment where multiple flares are added without the brightness adjustment operation.
In the experimental results, the addition of multiple flares increases the PSNR by less than $0.1$ dB, while the brightness adjustment operation leads to an increase of approximately $0.3$ dB.
These results indicate that simply adding multiple flares results in limited performance improvement and the primary enhancement is attributed to our brightness adjustment operation.

\par
\begin{table}[t]
	\centering
 	\caption{Ablation study of training network with different data synthesis methods. 
        "\ding{51}" indicates whether a certain effect acts on data synthesis.}
        \resizebox{0.95\textwidth}{!}{\begin{tabular}{c c @{\hspace{0.3cm}} c @{\hspace{0.3cm}} c c c c c}
            \toprule
      \multirow{2}{*}{Dataset} & \multirow{1}{*}{Adding} & \multirow{1}{*}{Brightness} & \multirow{2}{*}{PSNR$\uparrow$} & \multirow{2}{*}{SSIM$\uparrow$} & \multirow{2}{*}{LPIPS$\downarrow$} & \multirow{2}{*}{G-PSNR$\uparrow$} & \multirow{2}{*}{S-PSNR$\uparrow$} \\ 
     
     \multirow{1}{*}{}&\multirow{1}{*}{multiple flares} & \multirow{1}{*}{adjustment} \\
     \midrule
        Flare7K&\ding{55} & \ding{55} & 27.419 & 0.888 & 0.0459 & 23.788 & 21.111 \\ 
        - &\ding{51} & \ding{55} &  \underline{27.478} & \textbf{0.895} & \underline{0.0440} & \underline{23.833} & \underline{22.147} \\ 
        \textbf{Flare7K*} &\ding{51} & \ding{51} & \textbf{27.769} & \textbf{0.895} & \textbf{0.0429} & \textbf{24.010} & \textbf{22.849} \\ 
        \bottomrule
            \end{tabular}}
            \label{tab:data-sys}
            \vspace{-4mm}
\end{table}
%
%
The comparisons and analyses above demonstrate Flare7K*'s ability to enhance the model's performance in removing flares. 
Afterwards, we will apply our data synthesis method to the Flare7K++ dataset\cite{dai2023flare7k++} and train multiple models to test the universality of our synthesis method.
\vspace{-4mm}
\subsubsection{Adaptive Focus Module (AFM).} 
%
%
To address the issue of incorrectly touching the clean background, and leveraging the property that flare regions in images are brighter, we propose AFM.
This module enables the model to focus adaptively on local degradation, thus reducing the computational burden on the network.
We conduct a visual quality comparison between Uformer with and without AFM.
As shown in the images on the left of \cref{fig:data_sys}, we observe that models equipped with AFM demonstrate better performance in handling images with large areas of similar colours.
According to our analysis, the mask operation of AFM can remove low-light regions with colours similar to flares, effectively guiding the model to better detect and remove nighttime flares.
Besides, the model equipped with AFM does not perform incorrect background processing. As shown in the images on the right of \cref{fig:data_sys}, Uformer without AFM incorrectly removes shadows on the wall that resemble flares, while Uformer equipped with AFM does not.

\par

%
%
\begin{wraptable}{r}{6.4cm}
	\centering
 \vspace{-12.5mm}
	\caption{Ablation study of training with and without Adaptive Focus Module.}
	\resizebox{0.51\textwidth}{!}{\begin{tabular}{c c c | c@{\hspace{0.3cm}} c@{\hspace{0.3cm}} c@{\hspace{0.2cm}} c@{\hspace{0.2cm}} c}
            \toprule
            \multicolumn{3}{c}{AFM}& PSNR$\uparrow$   & SSIM$\uparrow$  & LPIPS$\downarrow$  & G-PSNR$\uparrow$ & S-PSNR$\uparrow$ \\ \midrule
\multicolumn{3}{c}{\ding{55}}& \underline{27.446} & \underline{0.890} & \underline{0.0461} & \underline{23.921} & \underline{22.706} 
\vspace{1.5mm}
\\
\multicolumn{3}{c}{\ding{51}}& \textbf{27.769} & \textbf{0.895} & \textbf{0.0429} & \textbf{24.010} & \textbf{22.849} \\ \bottomrule
            \end{tabular}}
            \vspace{-5mm}
            \label{tab:AFM}
\end{wraptable}
Additionally, we compare the quantitative results of the Uformer with and without AFM.
According to the results in \cref{tab:AFM}, the Uformer equipped with AFM achieves a nearly $0.323$ dB improvement in PSNR compared to the version without AFM.
We observe the effectiveness of AFM in conjunction with Uformer. Given its plug-and-play nature, we intend to validate its universality across multiple baselines to further demonstrate its broad applicability.
\par
\begin{figure*}[t]
\scriptsize
\centering
\begin{tabular}{cc}
\begin{adjustbox}{valign=t}
\begin{tabular}{cc} 
\includegraphics[width=0.46\textwidth]{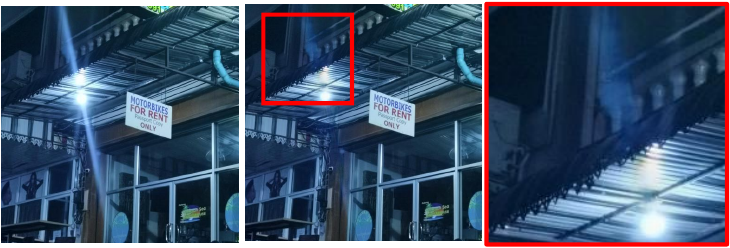}
\hspace{5mm}
\includegraphics[width=0.46\textwidth]{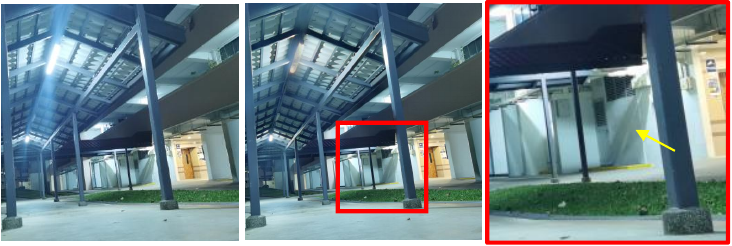}
\\
\hspace{-18mm}
(a) Input \hspace{4mm}
(b) w/o AFM \hspace{27mm}
(a) Input \hspace{4mm}
(b) w/o AFM
\\
\includegraphics[width=0.46\textwidth]{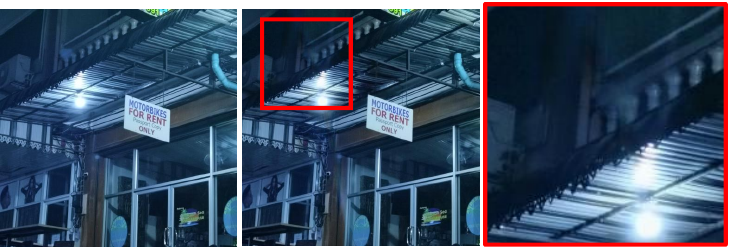}
\hspace{5mm}
\includegraphics[width=0.46\textwidth]{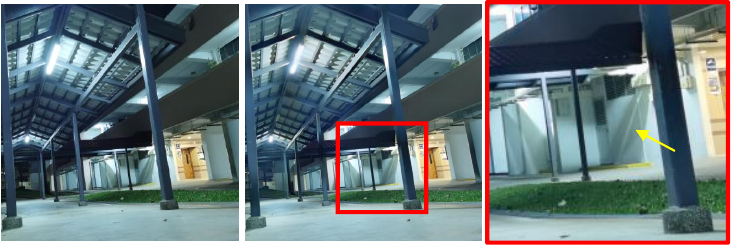}
\\
\hspace{-21mm}
(c) Reference   \hspace{2mm}
(d) w/ AFM      \hspace{27mm}
(c) Reference   \hspace{1.7mm}
(d) w/ AFM 
\end{tabular}
\end{adjustbox}
\end{tabular}
\caption{Visual comparison of the ablation experiment on Adaptive Focus Module.}
\label{fig:AFM}
\vspace{-7mm}
\end{figure*}
\vspace{-4mm}
\subsubsection{Adapt to Different Cameras.}
Given that different lenses are used in different scenarios, models trained on the same dataset may have limited generalization. 
By incorporating the field of view as a dynamic parameter in our data synthesis method, Flare7K* can be adapted to various cameras, offering a novel solution to address this limitation.
We train Uformer on different Flare7K* synthesized with varying fields of view and subsequently evaluate the performance on the real-world dataset.
\par
\begin{wrapfigure}{r}{6cm}
    \vspace{-10mm}
    \centering
    \includegraphics[width=0.5\textwidth]{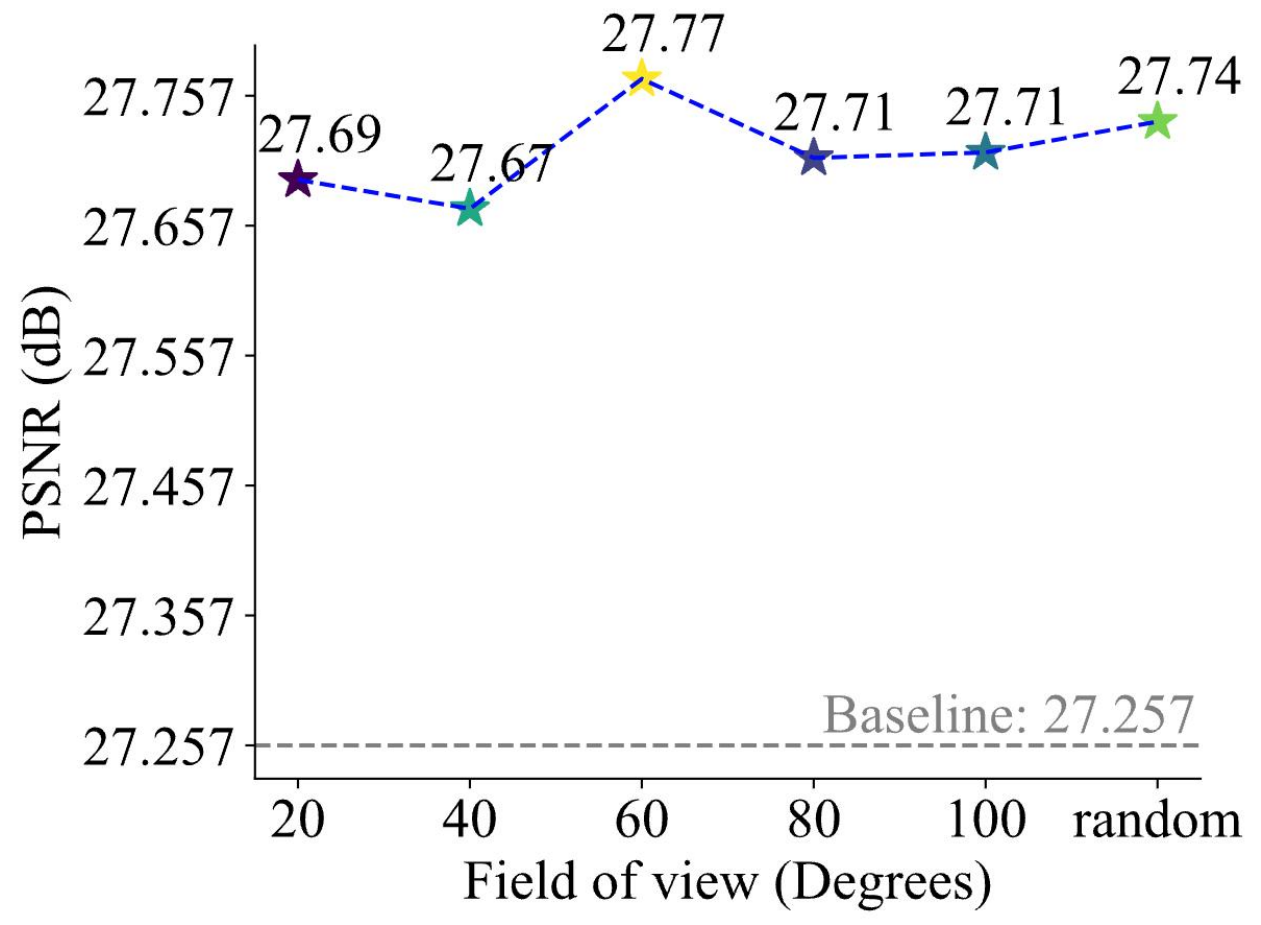}
    \vspace{-8mm}
    \caption{PSNR vs. field of view on the Flare7K\cite{dai2022flare7k} test dataset.}
    \label{fig:camera}
    \vspace{-7mm}
\end{wrapfigure}

As shown in \cref{fig:camera}, the difference between the highest and lowest does not exceed $0.1$ dB. 
This result can be attributed to the insensitivity of the field of view parameter in our method, and the inclusion of images from different cameras with varying fields of view in the test dataset.
The first factor indicates that our synthesis method is effective, rather than the improvement being solely due to a specific parameter.
Regarding the second factor, when photos are all taken with a specific camera, we believe that synthesizing the dataset using its field of view can yield superior results, which has significant implications for the industry.

%
\subsection{Extension to Other Baselines}
Given that our approach can easily be applied to existing networks, we use our data synthesis method and add AFM to these networks, including Restormer\cite{Zamir2021Restormer}, MPRNet\cite{MPRnet}, HINet\cite{chen2021hinet}, Uformer\cite{Uformer}.
Since these models were all trained on Flare7K++\cite{dai2023flare7k++} in the previous work of Dai \etal\cite{dai2023flare7k++}, we also train them on Flare7K*++. "*" denotes that it is synthesized by our proposed method.
\par
As shown in Table \ref{tab:other}, the PSNR of Restormer reaches $28.219$ dB, an improvement of nearly $0.59$ dB compared to the previous best of $27.633$ dB.
All metrics of these baselines are improved, which demonstrates the effectiveness and universality of our method.%
%


\begin{table*}[t]
\centering
\begin{center}
\caption{We enhance the models used in \cite{dai2023flare7k++} by incorporating our data synthesis method and AFM. For a fair comparison, all methods are trained on Flare7K++. The symbol "$\dag$" denotes that the model is the version with reduced parameters.}
\label{tab:other}
\setlength{\tabcolsep}{4pt}
\resizebox{0.99\textwidth}{!}{
\begin{tabular}{l c c c c  c c }
\toprule
Method & \multicolumn{1}{l}{Our technique}&\multicolumn{1}{c}{PSNR$\uparrow$} & \multicolumn{1}{c}{SSIM$\uparrow$} & \multicolumn{1}{c}{LPIPS$\downarrow$} & \multicolumn{1}{c}{G-PSNR$\uparrow$}& \multicolumn{1}{c}{S-PSNR$\uparrow$} \\
\hline
\multirow{2}{*}{Uformer \cite{Uformer}} 
& \ding{55} & 27.633 & 0.894 &0.0428 &23.949 & 22.603  \\
& \ding{51}& \underline{27.960} & \underline{0.899} & \textbf{0.0420} &24.249 &  \textbf{23.513}  \\
\hline
\multirow{2}{*}{HINet \cite{chen2021hinet}} 
& \ding{55} & 27.548 &0.892 &0.0464 &24.081 & 22.907  \\
& \ding{51}& 27.749&  0.892 &0.0435 &\underline{24.360} &  \underline{23.436}  \\
\hline
\multirow{2}{*}{$\dag$MPRNet \cite{MPRnet}} 
& \ding{55} & 27.036 &0.893&0.0481 &23.490 & 22.267  \\
& \ding{51}& 27.282 &0.896 &0.0474 &24.235 &  23.147   \\
\hline
\multirow{2}{*}{$\dag$Restormer\cite{Zamir2021Restormer}} 
& \ding{55} & 27.597 & 0.897 &0.0447 &23.828 & 22.452  \\
& \ding{51}& \textbf{28.219} &\textbf{0.902} &\underline{0.0427} &\textbf{24.569} &23.270   \\
\bottomrule
\end{tabular}}
\vspace{-5mm}
\end{center}
\end{table*}
\section{Conclusion}
To address the challenge of simulating a variety of scenes in the real world, we introduce a physically realistic data synthesis method based on the laws of illumination. Additionally, we develop an Adaptive Focus Module (AFM) which can help several models achieve better flare removal performance and avoid incorrectly modifying the clean background.
Most importantly, our data synthesis method can be applied to other flare removal datasets, and the AFM can be seamlessly integrated into various models, significantly improving their performance in nighttime flare removal.
Extensive comparative experiments and ablation studies, conducted across a range of scenarios, robustly demonstrate the effectiveness of our method.

\subsubsection{Limitations.}
Like previous works, we utilize the 24K Flickr image dataset\cite{zhang2018single} as the background image. However, since it does not include nighttime shots, it may not adapt well to nighttime scenes. This limitation can be addressed by incorporating clean nighttime background images.

%
%
%
\bibliographystyle{splncs04}
\bibliography{main}
\end{document}